%% file: Progressive_Transformers_for_E2E_Sign_Language_Production.tex
\documentclass[runningheads]{llncs}
\usepackage{graphicx}
\usepackage{comment}
\usepackage{amsmath,amssymb}
\usepackage{color}

\usepackage[nolist]{acronym}
\input{acronyms}
\usepackage{soul}
\usepackage{todonotes}
\usepackage{booktabs}
\usepackage{tablefootnote}
\usepackage{microtype}
\def\B{\fontseries{b}\selectfont}
\usepackage{hyperref}
\hypersetup{
    colorlinks=true,
    linkcolor=blue,
    filecolor=magenta,      
    urlcolor=cyan,
}

\begin{document}

\pagestyle{headings}
\mainmatter

\title{Progressive Transformers for End-to-End Sign Language Production}
\titlerunning{Progressive Transformers for End-to-End SLP}
\author{Ben Saunders\and
Necati Cihan Camgoz \and
Richard Bowden}
\authorrunning{B. Saunders et al.}
\institute{University of Surrey \\
\email{\{b.saunders,n.camgoz,r.bowden\}@surrey.ac.uk}}

\maketitle





\begin{abstract}

The goal of automatic \ac{slp} is to translate spoken language to a continuous stream of sign language video at a level comparable to a human translator. If this was achievable, then it would revolutionise Deaf hearing communications. Previous work on predominantly isolated \ac{slp} has shown the need for architectures that are better suited to the continuous domain of full sign sequences. 

In this paper, we propose Progressive Transformers, the first \ac{slp} model to translate from discrete spoken language sentences to continuous 3D sign pose sequences in an end-to-end manner. A novel counter decoding technique is introduced, that enables continuous sequence generation at training and inference. We present two model configurations, an end-to-end network that produces sign direct from text and a stacked network that utilises a gloss intermediary. We also provide several data augmentation processes to overcome the problem of drift and drastically improve the performance of \ac{slp} models. 

We propose a back translation evaluation mechanism for \ac{slp}, presenting benchmark quantitative results on the challenging \acf{ph14t} dataset and setting baselines for future research. Code available at \url{https://github.com/BenSaunders27/ProgressiveTransformersSLP}.

\end{abstract}

\section{Introduction}
Sign language is the language of communication for the Deaf community, a rich visual language with complex grammatical structures.  As it is their native language, most Deaf people prefer using sign as their main medium of communication, as opposed to a written form of spoken language. \acf{slp}, converting spoken language to continuous sign sequences, is therefore essential in involving the Deaf in the predominantly spoken language of the wider world. Previous work has been limited to the production of concatenated isolated signs \cite{stoll2018sign,zelinka2019nn}, highlighting the need for improved architectures to properly address the full remit of continuous sign language.
\begin{figure}[t!]
    \centering
    \includegraphics[width=0.74\linewidth]{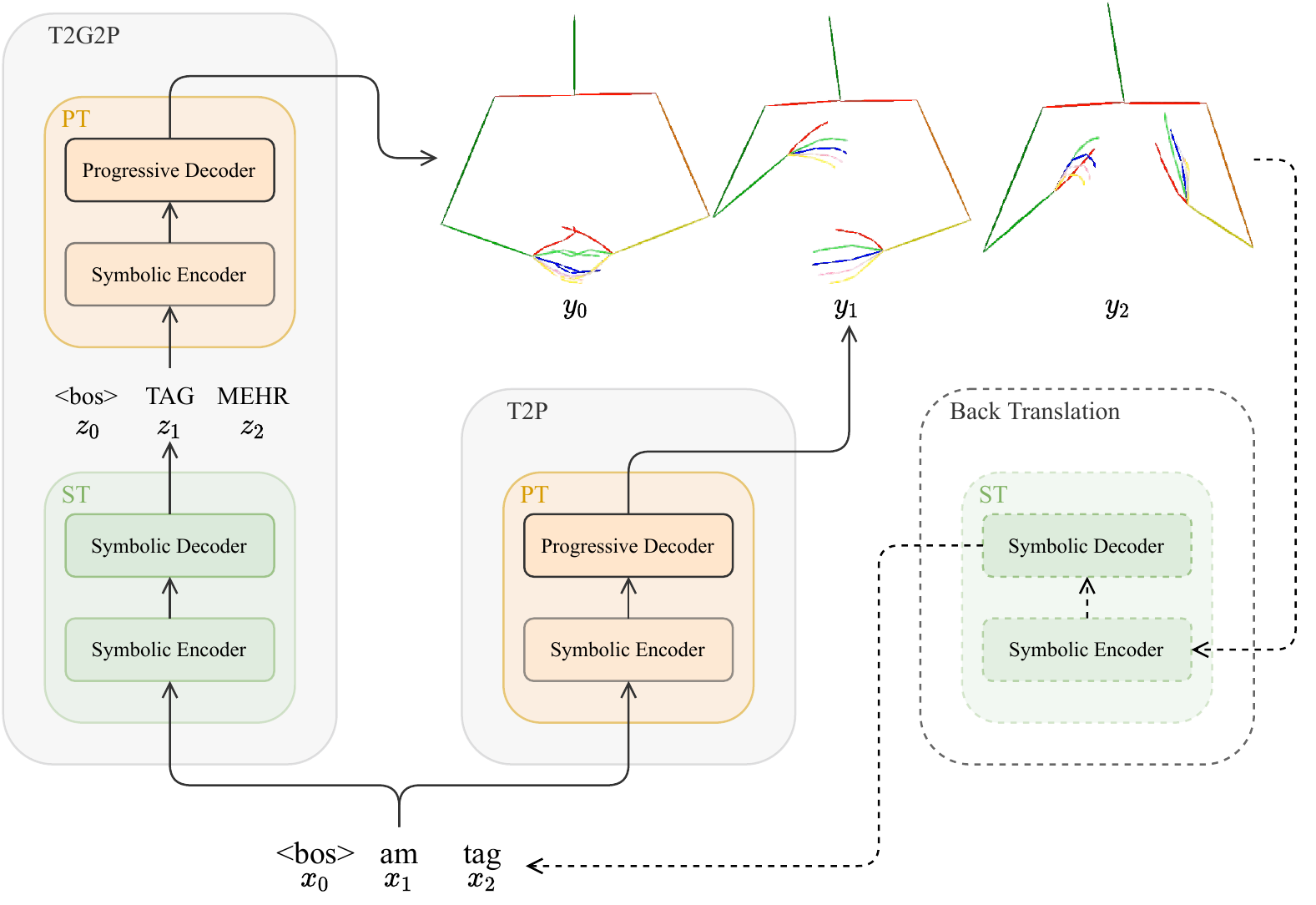}
    \caption{Overview of the Progressive Transformer architecture, showing Text to Gloss to Pose (T2G2P) and Text to Pose (T2P) model configurations. (PT: Progressive Transformer, ST: Symbolic Transformer)}
    \label{fig:architecture_overview}
\end{figure}

In this paper, we propose \textit{Progressive Transformers}, the first \ac{slp} model to translate from text to continuous 3D sign pose sequences in an end-to-end manner. Our novelties include an alternative formulation of transformer decoding for continuous variable sequences, where there is no pre-defined vocabulary. We introduce a counter decoding technique to predict continuous sequences of variable lengths by tracking the production progress, hence the name \textit{Progressive Transformers}. This approach also enables the driving of timing at inference, producing stable sign pose outputs. We also propose several data augmentation methods that assist in reducing drift in model production.

An overview  of our approach is shown in Figure \ref{fig:architecture_overview}. We evaluate two different model configurations, first translating from spoken language to sign pose via gloss\footnote{Glosses are a written representation of sign, defined as minimal lexical items.} intermediary (T2G2P), as this has been shown to increase translation performance \cite{camgoz2018neural}. In the second configuration we go direct, translating end-to-end from spoken language to sign (T2P).

To evaluate performance, we propose a back translation evaluation method for \ac{slp}, using a \ac{slt} model to translate back to spoken language (dashed lines in Figure \ref{fig:architecture_overview}). We evaluate on the challenging \acf{ph14t} dataset, presenting several benchmark results to underpin future research. We also share qualitative results to give further insight of the models performance to the reader, producing accurate sign pose sequences of an unseen text sentence.

The rest of this paper is organised as follows: In Section~\ref{sec:related_work}, we go over the previous research on \ac{slt} and \ac{slp}. In Section~\ref{sec:methodology}, we introduce our Progressive Transformer \ac{slp} model. Section~\ref{sec:quant_experiments} outlines the evaluation protocol and presents quantitative results, whilst Section \ref{sec:qual_experiments} showcases qualitative examples. Finally, we conclude the paper in Section \ref{sec:conc} by discussing our findings and possible future work. \looseness=-1

\section{Related Work} \label{sec:related_work}
\subsubsection{Sign Language Recognition \& Translation:}
Sign language has been a focus of computer vision researchers for over 30 years \cite{bauer2000video,starner1997real,tamura1988recognition}, primarily on isolated \ac{slr} \cite{ozdemir2016isolated,suzgun2015hospisign} and, relatively recently, the more demanding task of \ac{cslr} \cite{koller2015continuous,camgoz2017subunets}. However, the majority of work has relied on manual feature representations \cite{cooper2012sign} and statistical temporal modelling \cite{vogler1999parallel}. The availability of larger datasets, such as \ac{ph14} \cite{forster2014extensions}, have enabled the application of deep learning approaches such as \acp{cnn} \cite{koller2019weakly,koller2016deephand,koller2016deepsign} and \acp{rnn} \mbox{\cite{cui2017recurrent,koller2017resign}}.

Distinct to \ac{slr}, the task of \ac{slt} was recently introduced by Camgoz et al. \cite{camgoz2018neural}, aiming to directly translate sign videos to spoken language sentences \cite{duarte2019cross,ko2019neural,orbay2020neural,yin2020sign}. \ac{slt} is more challenging than \ac{cslr} due to the differences in grammar and ordering between sign and spoken language. Transformer based models are the current state-of-the-art in \ac{slt}, jointly learning the recognition and translation tasks~\cite{camgoz2020sign}.
\subsubsection{Sign Language Production:}
Previous approaches to \ac{slp} have extensively used animated avatars \cite{glauert2006vanessa,karpouzis2007educational,mcdonald2016automated} that can generate realistic sign production, but rely on phrase lookup and pre-generated sequences. \ac{smt} has also been applied to \ac{slp} \cite{kayahan2019hybrid,kouremenos2018statistical}, relying on static rule-based processing that can be difficult to encode.

Recently, deep learning approaches have been applied to the task of \ac{slp} \cite{duarte2019cross,stoll2018sign,xiao2020skeleton}. Stoll et al. present an initial \ac{slp} model using a combination of \ac{nmt} and \acp{gan} \cite{stoll2020text2sign}. The authors break the problem into three separate processes that are trained independently, producing a concatenation of isolated 2D skeleton poses \cite{ebling2018smile} mapped from sign glosses via a look-up table. Contrary to Stoll et al., our paper focuses on automatic sign production and learning the mapping between text and skeleton pose sequences directly, instead of providing this a priori.

The closest work to this paper is that of Zelinka et al., who build a neural-network-based translator between text and synthesised skeletal pose \cite{zelinka2020neural}. The authors produce a single sign for each source word with a set size of 7 frames, generating sequences with a fixed length and ordering. In contrast, our model allows a dynamic length of output sign sequence, learning the correct length and ordering of each word from the data, whilst using counter decoding to determine the end of sequence generation. Unlike \cite{zelinka2020neural}, who work on a proprietary dataset, we produce results on the publicly available \ac{ph14t}, providing a benchmark for future \ac{slp} research.
\subsubsection{Neural Machine Translation:}
\ac{nmt} aims to learn a mapping between language sequences, generating a target sequence from a source sequence of another language. \acp{rnn} were first proposed to solve the sequence-to-sequence problem, with Kalchbrenner et al. \cite{kalchbrenner2013recurrent} introducing a single \ac{rnn} that iteratively applied a hidden state computation. Further models were later developed \cite{cho2014properties,sutskever2014sequence} that introduced encoder-decoder architectures, mapping both sequences to an intermediate embedding space. Bahdanau et al. \cite{bahdanau2014neural} overcame the bottleneck problem by adding an attention mechanism that facilitated a soft-search over the source sentence for the context most useful to the target word prediction.

Transformer networks \cite{vaswani2017attention}, a recent \ac{nmt} breakthrough, are based solely on attention mechanisms, generating a representation of the entire source sequence with global dependencies. \ac{mha} is used to model different weighted combinations of an input sequence, improving the representation power of the model. Transformers have achieved impressive results in many classic \ac{nlp} tasks such as language modelling \cite{dai2019transformer,zhang2019ernie} and sentence representation \cite{devlin2018bert} alongside other domains including image captioning \cite{li2019entangled,zhou2018end} and action recognition \cite{girdhar2019video}. Related to this work, transformer networks have previously been applied to continuous output tasks such as speech synthesis \cite{li2019neural,ren2019fastspeech,vila2018end}, music production \cite{huang2018music} and image generation \cite{parmar2018image}.

Applying \ac{nmt} methods to continuous output tasks is a relatively underresearched problem. Encoder-decoder models and \acp{rnn} have been used to map text to a human action sequence \cite{ahn2018text2action,plappert2018learning} whilst adversarial discriminators have enabled the production of realistic pose \cite{ginosar2019learning,lee2019dancing}. In order to determine sequence length of continuous outputs, previous works have used a fixed output size that limits the models flexibility \cite{zelinka2020neural}, a binary end-of-sequence (EOS) flag \cite{graves2013generating} or a continuous representation of an EOS token \cite{mukherjee2019predicting}.

\section{Progressive Transformers} \label{sec:methodology}
In this section, we introduce \textit{Progressive Transformers}, an \ac{slp} model which learns to translate spoken language sentences to continuous sign pose sequences. Our objective is to learn the conditional probability $p(Y|X)$ of producing a sequence of signs $Y = (y_{1},...,y_{U})$ with $U$ time steps, given a spoken language sentence $X = (x_{1},...,x_{T})$ with $T$ words. Gloss can also be used as intermediary supervision for the network, formulated as $Z = (z_{1},...,z_{N})$ with $N$ glosses, where the objective is then to learn the conditional probabilities $p(Z|X)$ and $p(Y|Z)$.

Producing a target sign sequence from a reference text sequence poses several challenges. Firstly, the sequences have drastically varying length, with the number of frames much larger than the number of words ($U >> T$). The sequences also have a non-monotonic relationship due to the different vocabulary and grammar used in sign and spoken languages. Finally, the target signs inhabit a continuous vector space requiring a differing representation to the discrete space of text.

\begin{figure}[t!]
    \centering
    \includegraphics[width=1.0 \linewidth]{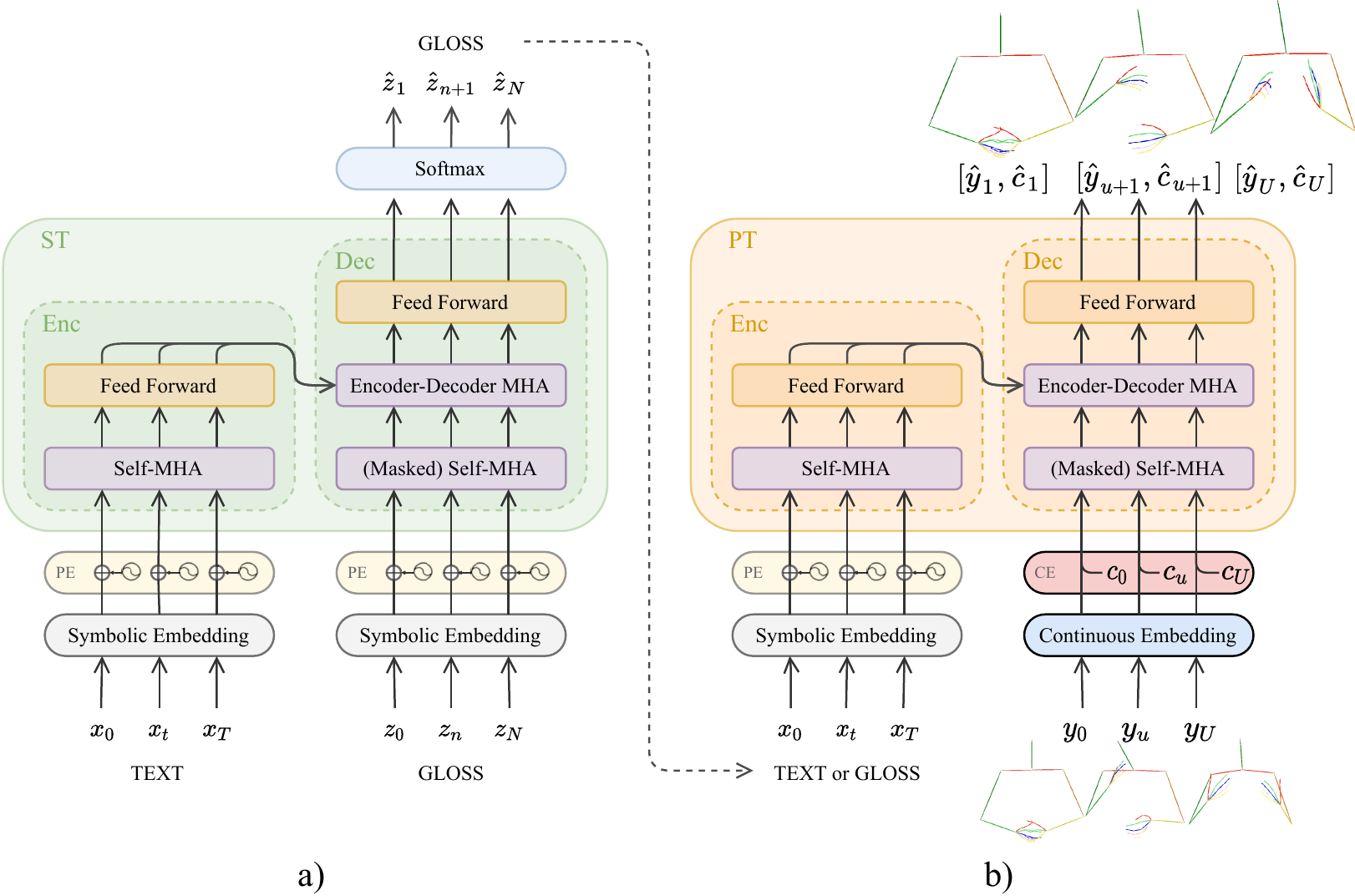}
    \caption{Architecture details of (a) Symbolic and (b) Progressive Transformers. \protect\linebreak (ST: Symbolic Transformer, PT: Progressive Transformer, PE: Positional Encoding, CE: Counter Embedding, MHA: Multi-Head Attention)}
    \label{fig:architecture_details}
\end{figure}

To address the production of continuous sign sequences, we propose a progressive transformer-based architecture that allows translation from a symbolic to a continuous sequence domain. We first formalise a Symbolic Transformer architecture, converting an input to a symbolic target feature space, as detailed in Figure \ref{fig:architecture_details}a. This is used in our \acf{ttgtp} model to convert from spoken language to gloss representation as an intermediary step before pose production, as seen in Figure \ref{fig:architecture_overview}.

We then describe the Progressive Transformer architecture, translating from a symbolic input to a continuous output representation, as shown in Figure \ref{fig:architecture_details}b. We use this model for the production of realistic and understandable sign language sequences, either via gloss supervision in the \ac{ttgtp} model or direct from spoken language in our end-to-end \acf{ttp} model. To enable sequence length prediction of a continuous output, we introduce a counter decoding that allows the model to track the progress of sequence generation. In the remainder of this section we describe each component of the architecture in detail.
\subsection{Symbolic Transformer}
We build on the classic transformer \cite{vaswani2017attention}, a model designed to learn the mapping between symbolic source and target languages. In this work, Symbolic Transformers (Figure \ref{fig:architecture_details}a) translate from source text to target gloss sequences. As per the standard \ac{nmt} pipeline \cite{mikolov2013distributed}, we first embed the source, $x_{t}$, and target, $z_{n}$, tokens via a linear embedding layer, to represent the one-hot-vector in a higher-dimensional space where tokens with similar meanings are closer. Symbolic embedding, with weight, $W$, and bias, $b$, can be formulated as:
\begin{equation}
\label{eq:word_embedding}
    w_{t} = W^{x} \cdot x_{t} + b^{x},\;\;\;\;  g_{n} = W^{z} \cdot z_{n} + b^{z}
\end{equation}
where $w_{t}$ and $g_{n}$ are the vector representations of the source and target tokens.

Transformer networks do not have a notion of word order, as all source tokens are fed to the network simultaneously without positional information. To compensate for this and provide temporal ordering, we apply a temporal embedding layer after each input embedding. For the symbolic transformer, we apply positional encoding \cite{vaswani2017attention}, as:
\begin{equation}
\label{eq:word_PE}
    \hat{w}_{t} = w_{t} + \textrm{PositionalEncoding}(t)
\end{equation}
\begin{equation}
\label{eq:gloss_PE}
    \hat{g}_{n} = g_{n} + \textrm{PositionalEncoding}(n)
\end{equation}
where PositionalEncoding is a predefined sinusoidal function conditioned on the relative sequence position $t$ or $n$. 

Our symbolic transformer model consists of an encoder-decoder architecture. The encoder first learns the contextual representation of the source sequence through self-attention mechanisms, understanding each input token in relation to the full sequence. The decoder then determines the mapping between the source and target sequences, aligning the representation sub-spaces and generating target predictions in an auto-regressive manner.

The symbolic encoder ($E_{S}$) consists of a stack of $L$ identical layers, each containing 2 sub-layers. Given the temporally encoded source embeddings, $\hat{w}_{t}$, a \ac{mha} mechanism first generates a weighted contextual representation, performing multiple projections of scaled dot-product attention. This aims to learn the relationship between each token of the sequence and how relevant each time step is in the context of the full sequence. Formally, scaled dot-product attention outputs a vector combination of values, $V$, weighted by the relevant queries, $Q$, keys, $K$, and dimensionality, $d_{k}$:
\begin{equation}
\label{eq:attention}
    \textrm{Attention}(Q,K,V) = \text{softmax}(\frac{Q K^{T}}{\sqrt{d_{k}}})V
\end{equation}

\ac{mha} stacks parallel attention mechanisms in $h$ different mappings of the same queries, keys and values, each with varied learnt parameters. This allows different representations of the input to be generated, learning complementary information in different sub-spaces. The outputs of each head are then concatenated together and projected forward via a final linear layer, as:
\begin{align}
\label{eq:multi_head_attention}
    \textrm{MHA}(Q,K, & V)  =  [head_{1}, ... ,head_{h}] \cdot W^{O}, \nonumber \\
      &  \textrm{where} \medspace head_{i} = \textrm{Attention}(QW_{i}^{Q}, KW_{i}^{K}, VW_{i}^{V})
\end{align}
and $W^{O}$,$W_{i}^{Q}$,$W_{i}^{K}$ and $W_{i}^{V}$ are weights related to each input variable.

The outputs of \ac{mha} are then fed into the second sub-layer of a non-linear feed-forward projection. A residual connection \cite{he2016deep} and subsequent layer norm \cite{ba2016layer} is employed around each of the sub-layers, to aid training. The final symbolic encoder output can be formulated as:
\begin{equation}
\label{eq:symbolic_encoder}
    h_{t} = E_{S}(\hat{w}_{t}  | \hat{w}_{1:T})
\end{equation}
where $h_{t}$ is the contextual representation of the source sequence.

The symbolic decoder ($D_{S}$) is an auto-regressive architecture that produces a single token at each time-step. The positionally embedded target sequences, $\hat{g}_{n}$, are passed through an initial MHA self-attention layer similar to the encoder, with an extra masking operation. Alongside the fact that the targets are offset from the inputs by one position, the masking of future frames prevents the model from attending to subsequent time steps in the sequence.

A further \ac{mha} sub-layer is then applied, which combines encoder and decoder representations and learns the alignment between the source and target sequences. The final sub-layer is a feed forward layer, as in the encoder. After all decoder layers are processed, a final non-linear feed forward layer is applied, with a softmax operation to generate the most likely output token at each time step. The output of the symbolic decoder can be formulated as:
\begin{equation}
\label{eq:symbolic_decoder}
    z_{n+1} = \operatorname*{argmax}_{i} D_{S}(\hat{g}_{n}  | \hat{g}_{1:n-1} , h_{1:T} )
\end{equation}
where $z_{n+1}$ is the output at time $n+1$, from a target vocabulary of size $i$.

\subsection{Progressive Transformer}
We now adapt our symbolic transformer architecture to cope with continuous outputs, in order to convert source sequences to a continuous target domain. In this work, Progressive Transformers (Figure \ref{fig:architecture_details}b) translate from the symbolic domains of gloss or text to continuous sign pose sequences that represent the motion of a signer producing a sentence of sign language. The model must produce skeleton pose outputs that can both express an accurate translation of the given input sequence and a realistic sign pose sequence. 

We represent each sign pose frame, $y_{u}$, as a continuous vector of the 3D joint positions of the signer. These joint values are first passed through a linear embedding layer, allowing sign poses of similar content to be closely represented in the dense space. The continuous embedding layer can be formulated as:
\begin{equation}
\label{eq:joint_embedding}
    j_{u} = W^{y} \cdot y_{u} + b^{y}
\end{equation}
where $j_{u}$ is the embedded 3D joint coordinates of each frame, $y_{u}$.

We next apply a counter embedding layer to the sign poses as temporal embedding (CE in Figure \ref{fig:architecture_details}). The counter, $c$, holds a value between 0 and 1, representing the frame position relative to the total sequence length. The joint embeddings, $j_{u}$, are concatenated with the respective counter value, $c_{u}$, formulated as:
\begin{equation}
\label{eq:counter_appending}
    \hat{j}_{u} = [j_{u},\textrm{CounterEmbedding}(u)]
\end{equation}
where CounterEmbedding is a linear projection of the counter value for frame $u$. 

\begin{figure}[t!]
    \centering
    \includegraphics[width=0.95 \linewidth]{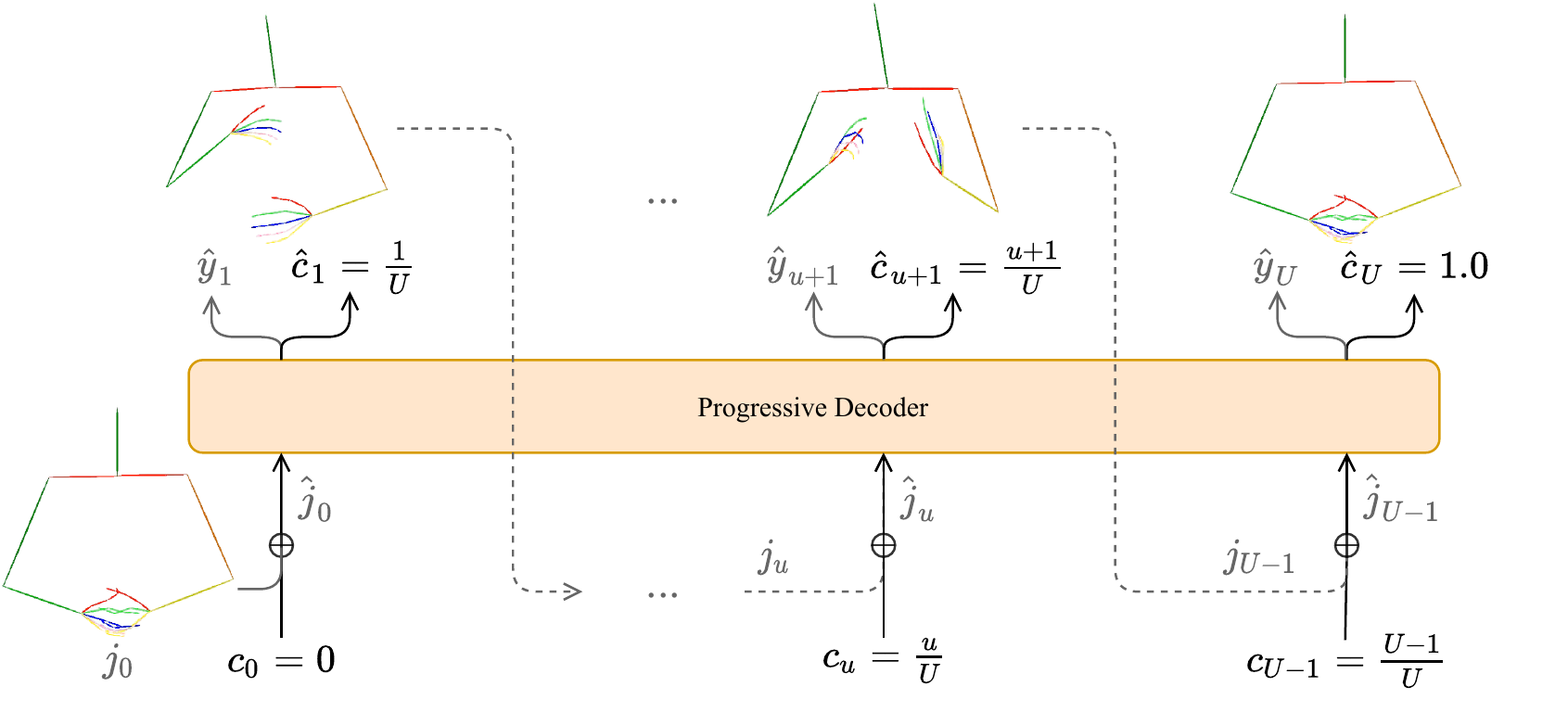}
    \caption{Counter decoding example, showing the simultaneous prediction of sign pose, $\hat{y}_{u}$, and counter, $\hat{c}_{u} \in \{0:1\}$, with $\hat{c}=1.0$ denoting end of sequence}
    \label{fig:counter_concatenation}
\end{figure}

At each time-step, counter values are predicted alongside the skeleton pose, as shown in Figure \ref{fig:counter_concatenation}, with sequence generation concluded once the counter reaches 1. We call this process Counter Decoding, determining the progress of sequence generation and providing a way to predict the end of sequence without the use of a tokenised vocabulary.

The counter provides the model with information relating to the length and speed of each sign pose sequence, determining the sign duration. At inference, we drive the sequence generation by replacing the predicted counter value, $\hat{c}$, with the ground truth timing information, $c^{*}$, to produce a stable output sequence.

The Progressive Transformer also consists of an encoder-decoder architecture. Due to the input coming from a symbolic source, the encoder has a similar setup to the symbolic transformer, learning a contextual representation of the input sequence. As the representation will ultimately be used for the end goal of \ac{slp}, these representations must also contain sufficient context to fully and accurately reproduce sign. Taking as input the temporally embedded source embeddings, $\hat{w}_{t}$, the encoder can be formulated as:
\begin{equation}
\label{eq:progressive_encoder}
    r_{t} = E_{S} (\hat{w}_{t}  | \hat{w}_{1:T})
\end{equation}
where $E_{S}$ is the symbolic encoder and $r_{t}$ is the encoded source representation.

The progressive decoder ($D_{P}$) is an auto-regressive model that produces a sign pose frame at each time-step, alongside the counter value described above. Distinct from symbolic transformers, the progressive decoder produces continuous sequences that hold a sparse representation in a large continuous sub-space. The counter-concatenated joint embeddings, $\hat{j}_{u}$, are extracted as target input, representing the sign information of each frame. 

A self-attention \ac{mha} sub-layer is first applied, with target masking to avoid attending to future positions. A further \ac{mha} mechanism is then used to map the symbolic representations from the encoder to the continuous domain of the decoder, learning the important alignment between spoken and sign languages. 

A final feed forward sub-layer follows, with each sub-layer followed by a residual connection and layer normalisation as before. No softmax layer is used as the skeleton joint coordinates can be regressed directly and do not require stochastic prediction. The progressive decoder output can be formulated as:
\begin{equation}
\label{eq:progressive_decoder}
    [\hat{y}_{u+1},\hat{c}_{u+1}] = D_{P}(\hat{j}_{u}  | \hat{j}_{1:u-1} , r_{1:T} )
\end{equation}
where $\hat{y}_{u+1}$ corresponds to the 3D joint positions representing the produced sign pose of frame $u+1$ and $\hat{c}_{u+1}$ is the respective counter value. The decoder learns to generate one frame at a time until the predicted counter value reaches 1, determining the end of sequence. Once the full sign pose sequence is produced, the model is trained end-to-end using the \ac{mse} loss between the predicted sequence, $\hat{y}_{1:U}$, and the ground truth, $y_{1:U}^{*}$:
\begin{equation}
\label{eq:loss_mse}
    L_{MSE} = \frac{1}{U} \sum_{i=1}^{u} ( y_{1:U}^{*} - \hat{y}_{1:U} ) ^{2}
\end{equation}

The progressive transformer outputs, $\hat{y}_{1:U}$, represent the 3D skeleton joint positions of each frame of a produced sign sequence. To ease the visual comparison with reference sequences, we apply \ac{dtw} \cite{berndt1994dtw} to align the produced sign pose sequences. Animating a video from this sequence is then a trivial task, plotting the joints and connecting the relevant bones, with timing information provided from the counter. These 3D joints could subsequently be used to animate an avatar \cite{kipp2011sign,mcdonald2016automated} or condition a \ac{gan} \cite{isola2017image,zhu2017unpaired}.

\section{Quantitative Experiments} \label{sec:quant_experiments}
In this section, we share our \ac{slp} experimental setup and report experimental results. We first provide dataset and evaluation details, outlining back translation. We then evaluate both symbolic and progressive transformer models, demonstrating results of data augmentation and model configuration.
\subsection{Sign Language Production Dataset} \label{sec:dataset}

Forster et al. released \ac{ph14} \cite{forster2014extensions} as a large video-based corpus containing parallel sequences of \ac{dgs} and spoken text extracted from German weather forecast recordings. This dataset is ideal for computational sign language research due to the provision of gloss level annotations, becoming the primary benchmark for both \ac{slr} and \ac{cslr}.

In this work, we use the publicly available \ac{ph14t} dataset introduced by Camgoz et al. \cite{camgoz2018neural}, a continuous \ac{slt} extension of the original \ac{ph14}. This corpus includes parallel sign videos and German translation sequences with redefined segmentation boundaries generated using the forced alignment approach of \cite{koller2016deepsign}. 8257 videos of 9 different signers are provided, with a vocabulary of 2887 German words and 1066 different sign glosses from a combined 835,356 frames.

\begin{figure}[t!]
    \centering
    \includegraphics[width=0.70 \linewidth]{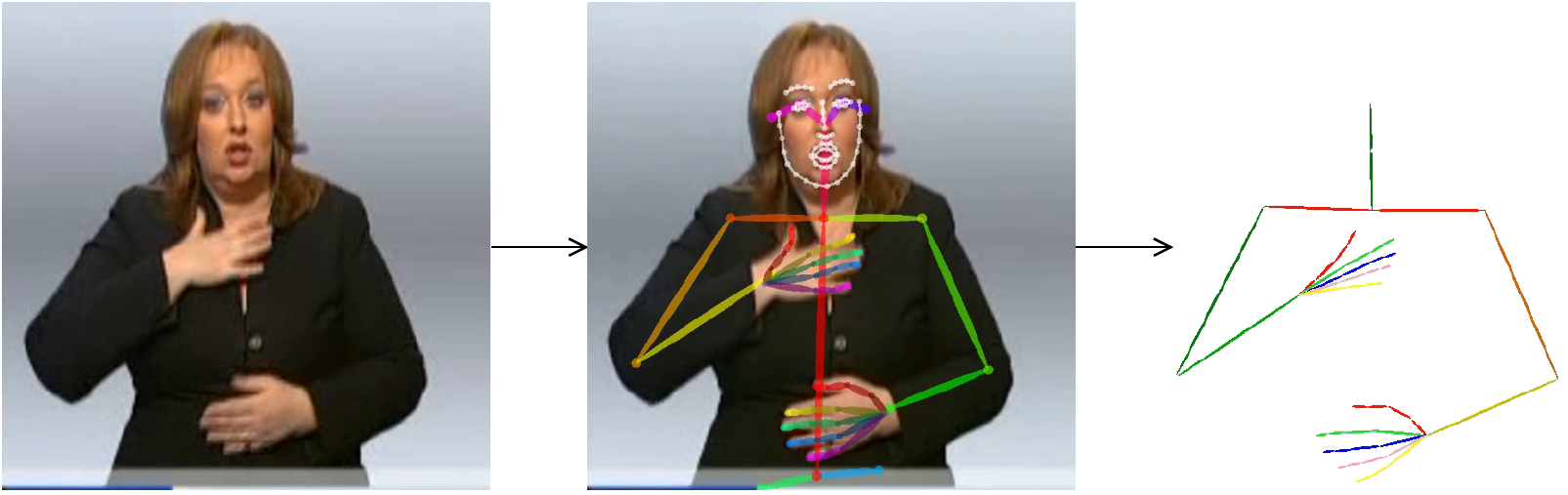}
    \caption{Skeleton pose extraction, using OpenPose \cite{cao2018openpose} and 2D to 3D mapping \cite{zelinka2020neural}}
    \label{fig:skel_example}
\end{figure}

We train our \ac{slp} network to generate sequences of 3D skeleton pose. 2D joint positions are first extracted from each video using OpenPose \cite{cao2018openpose}. We then utilise the skeletal model estimation improvements presented in \cite{zelinka2020neural} to lift the 2D joint positions to 3D. An iterative inverse kinematics approach is applied to minimise 3D pose whilst maintaining consistent bone length and correcting misplaced joints. Finally, we apply skeleton normalisation similar to \cite{stoll2018sign} and represent 3D joints as $x$, $y$ and $z$ coordinates. An example is shown in Figure \ref{fig:skel_example}.
\subsection{Evaluation Details} \label{sec:evaluation_details}
In this work, we propose back-translation as a means of \ac{slp} evaluation, translating back from produced sign to spoken language. This provides a measure of how understandable the productions are, and how much translation content is preserved. Evaluation of a generative model is often difficult but we find a close correspondence between back translation score and the visual production quality. We liken it to the wide use of the inception score for generative models \cite{salimans2016improved}, using a pre-trained classifier. Similarly, recent \ac{slp} work used an \ac{slr} discriminator to evaluate isolated skeletons \cite{xiao2020skeleton}, but did not measure the translation performance.

We utilise the state-of-the-art \ac{slt} \cite{camgoz2020sign} as our back translation model, modified to take sign pose sequences as input. This is again trained on the \ac{ph14t} dataset, ensuring a robust translation from sign to text. We generate spoken language translations of the produced sign pose sequences and compute BLEU and ROUGE scores. We provide BLEU n-grams from 1 to 4 for completeness.

In the following experiments, our symbolic and progressive transformer models are each built with 2 layers, 8 heads and embedding size of 256. All parts of our network are trained with Xavier initialisation \cite{glorot2010understanding}, Adam optimization \cite{kingma2014adam} with default parameters and a learning rate of $10^{-3}$. Our code is based on Kreutzer et al.'s NMT toolkit, JoeyNMT \cite{JoeyNMT}, and implemented using PyTorch \cite{paszke2017automatic}. 

\subsection{Symbolic Transformer: Text to Gloss}
Our first experiment measures the performance of the symbolic transformer architecture for sign language understanding. We train our symbolic transformer to predict gloss representations from source spoken language sentences. Table \ref{tab:text_to_gloss_results} shows our model achieves state-of-the-art results, significantly outperforming that of Stoll et al. \cite{stoll2018sign}, who use an encoder-decoder network with 4 layers of 1000 \acp{gru}. This supports our use of the proposed transformer architecture for sign language understanding.
\begin{table}[t!]
\centering
\caption{Symbolic Transformer results for Text to Gloss translation}
\resizebox{0.9\linewidth}{!}{%
\begin{tabular}{@{}p{2.8cm}ccccc|ccccc@{}}
\toprule
            & \multicolumn{5}{c}{DEV SET} & \multicolumn{5}{c}{TEST SET} \\ 
\multicolumn{1}{c|}{Approach:}      & BLEU-4         & BLEU-3         & BLEU-2         & BLEU-1        & ROUGE          & BLEU-4        & BLEU-3         & BLEU-2         & BLEU-1         & ROUGE          \\ \midrule
\multicolumn{1}{r|}{Stoll et al. \cite{stoll2018sign}} & 
16.34       & 22.30         & 32.47        & 50.15       & 48.42    & 15.26      & 21.54       & 32.25        & 50.67        & 48.10          \\
\multicolumn{1}{r|}{Ours}  &  {\B 20.23} & {\B 27.36} & {\B 38.21} & {\B 55.65} & {\B 55.41} & {\B 19.10} & {\B 26.24} & {\B 37.10} & {\B 55.18} & {\B 54.55} \\ \bottomrule
\end{tabular}%
}
\label{tab:text_to_gloss_results}
\end{table}
\subsection{Progressive Transformer: Gloss to Pose}
In our next set of experiments, we evaluate our progressive transformer and its capability to produce a continuous sign pose sequence from a given symbolic input. As a baseline, we train a progressive transformer model to translate from gloss to sign pose without augmentation, with results shown in Table \ref{tab:data_augmentation_results} (Base). 

We believe our base progressive model suffers from prediction drift, with erroneous predictions accumulating over time. As transformer models are trained to predict the next time-step of all ground truth inputs, they are often not robust to noise in target inputs. At inference time, with predictions based off previous outputs, errors are propagated throughout the full sequence generation, quickly leading to poor quality production. The impact of drift is heightened due to the continuous distribution of the target skeleton poses. As neighbouring frames differ little in content, a model learns to just copy the previous ground truth input and receive a small loss penalty. We thus experiment with various data augmentation approaches in order to overcome drift and improve performance.  
\begin{table}[t!]
\caption{Progressive Transformer results for Gloss to Sign Pose production, with multiple data augmentation techniques. FP: Future Prediction, GN: Gaussian Noise}
\centering
\resizebox{0.9\linewidth}{!}{%
\begin{tabular}{@{}p{2.8cm}ccccc|ccccc@{}}
\toprule
     & \multicolumn{5}{c}{DEV SET}  & \multicolumn{5}{c}{TEST SET} \\ 
\multicolumn{1}{c|}{Approach:}  & BLEU-4         & BLEU-3         & BLEU-2         & BLEU-1         & ROUGE          & BLEU-4         & BLEU-3         & BLEU-2         & BLEU-1         & ROUGE          \\ \midrule
\multicolumn{1}{r|}{Base}              & 7.04       & 9.10       & 13.12        & 24.20        & 25.53       & 5.03       & 6.89        & 10.81        & 23.03        & 23.31          \\
\multicolumn{1}{r|}{Future Prediction} & 9.96       & 12.71     & 17.83       & 30.03       & 31.03      & 8.38      & 11.04       & 16.41       & 28.94       & 29.73          \\
\multicolumn{1}{r|}{Just Counter}      & 11.04       & 13.86       & 19.05       & 31.16      & 32.45    & 9.16       & 11.96       & 17.41      & 30.08        & 30.41          \\
\multicolumn{1}{r|}{Gaussian Noise}    & 11.88 & 15.07 & {\B 20.61} & {\B 32.53} & {\B 34.19} &  10.02 & 12.96 & 18.58 & 31.11 & 31.83 \\ 
\multicolumn{1}{r|}{FP \& GN}    & {\B 11.93} & {\B 15.08} & 20.50 & 32.40 & 34.01 & {\B 10.43}  & {\B 13.51} & {\B 19.19} & {\B 31.80} & {\B 32.02} \\ \bottomrule
\end{tabular}%
}
\label{tab:data_augmentation_results}
\end{table}
\subsubsection{Future Prediction} 
Our first data augmentation method is conditional future prediction, requiring the model to predict more than just the next frame in the sequence. Experimentally, we find the best performance comes from a prediction of all of the next 10 frames from the current time step. As can be seen in Table \ref{tab:data_augmentation_results}, prediction of future time steps increases performance from the base architecture. We believe this is because the model now cannot rely on just copying the previous frame, as there are more considerable changes to the skeleton positions in 10 frames time. The underlying structure and movement of sign has to be learnt, encoding how each gloss is represented and reproduced in the training data. 
\subsubsection{Just Counter}
Inspired by the memorisation capabilities of transformer models, we next experiment with a pure memorisation approach. Only the counter values are provided as target input to the model, as opposed to the usual full 3D skeleton joint positions. We show a further performance increase with this approach, considerably increasing the BLEU-4 score as shown in Table \ref{tab:data_augmentation_results}.

We believe the just counter model setup helps to allay the effect of drift, as the model now must learn to decode the target sign pose solely from the counter position, without relying on the ground truth joint embeddings it previously had access to. This setup is now identical at both training and inference, with the model having to generalise only to new data rather than new prediction inputs.
\subsubsection{Gaussian Noise}
Our final augmentation experiment examines the effect of applying noise to the skeleton pose sequences during training, increasing the variety of data to train a more robust model. For each joint, statistics on the positional distribution of the previous epoch are collected, with randomly sampled noise applied to the inputs of the next epoch. Applied noise is multiplied by a noise factor, $r_{n}$, with empirical validation suggesting $r_{n} = 5$ gives the best performance. An increase of Gaussian noise causes the model to become more robust to prediction inputs, as it must learn to correct the augmented inputs back to the target outputs. 

Table \ref{tab:data_augmentation_results} (FP \& GN) shows that the best BLEU-4 performance comes from a combination of future prediction and Gaussian noise augmentation. The model must learn to cope with both multi-frame prediction and a noisy input, building a firm robustness to drift. We continue with this setup for further experiments.
\begin{table}[t!]
\centering
\caption{Results of the Text2Pose (T2P) and Text2Gloss2Pose (T2G2P) network configurations for Text to Sign Pose production}
\resizebox{0.9\linewidth}{!}{%
\begin{tabular}{@{}p{2.8cm}ccccc|ccccc@{}}
\toprule
 & \multicolumn{5}{c}{DEV SET} & \multicolumn{5}{c}{TEST SET} \\ 
\multicolumn{1}{c|}{Configuration:} & BLEU-4         & BLEU-3         & BLEU-2         & BLEU-1         & ROUGE          & BLEU-4         & BLEU-3         & BLEU-2         & BLEU-1         & ROUGE          \\ \midrule
\multicolumn{1}{r|}{T2P}     &  {\B 11.82} & {\B 14.80} & 19.97 & 31.41 & 33.18 & {\B 10.51} & {\B 13.54} & {\B 19.04} & {\B 31.36} & {\B 32.46} \\
\multicolumn{1}{r|}{T2G2P}   &  11.43 & 14.71 & {\B 20.71} & {\B 33.12} & {\B 34.05} & 9.68 & 12.53 & 17.62 & 29.74 & 31.07 \\ \bottomrule
\end{tabular}%
}
\label{tab:configuration_results}
\end{table}
\subsection{Text2Pose v Text2Gloss2Pose}
Our final experiment evaluates the two network configurations outlined in Figure \ref{fig:architecture_overview}, sign production either direct from text or via a gloss intermediary. Text to Pose (T2P) consists of a single progressive transformer model with spoken language input, learning to jointly translate from the domain of spoken language to sign and subsequently produce meaningful sign representations. Text to Gloss to Pose (T2G2P) uses an initial symbolic transformer to convert to gloss, which is then input into a further progressive transformer to produce sign pose sequences.

As can be seen from Table \ref{tab:configuration_results}, the T2P model outperforms that of T2G2P. This is surprising, as a large body of previous work has suggested that using gloss as intermediary helps networks learn \cite{camgoz2018neural}. However, we believe this is because there is more information available within spoken language compared to a gloss representation, with more tokens per sequence to predict from. Predicting gloss sequences as an intermediary can act as a bottleneck, as all information required for production needs to be present in the gloss. Therefore, any contextual information present in the source text can be lost.

The success of the T2P network shows that our progressive transformer model is powerful enough to complete two sub-tasks; firstly mapping spoken language sequences to a sign representation, then producing an accurate sign pose recreation. This is important for future scaling and application of the \ac{slp} model architecture, as many sign language domains do not have gloss availability. 

Furthermore, our final BLEU-4 scores outperform similar end-to-end Sign to Text methods which do not utilize gloss information \cite{camgoz2018neural} (9.94 BLEU-4). Note that this is an unfair direct comparison, but it does provide an indication of model performance and the quality of the produced sign pose sequences.
\begin{figure}[t!]
    \centering
    \includegraphics[width=0.95\linewidth]{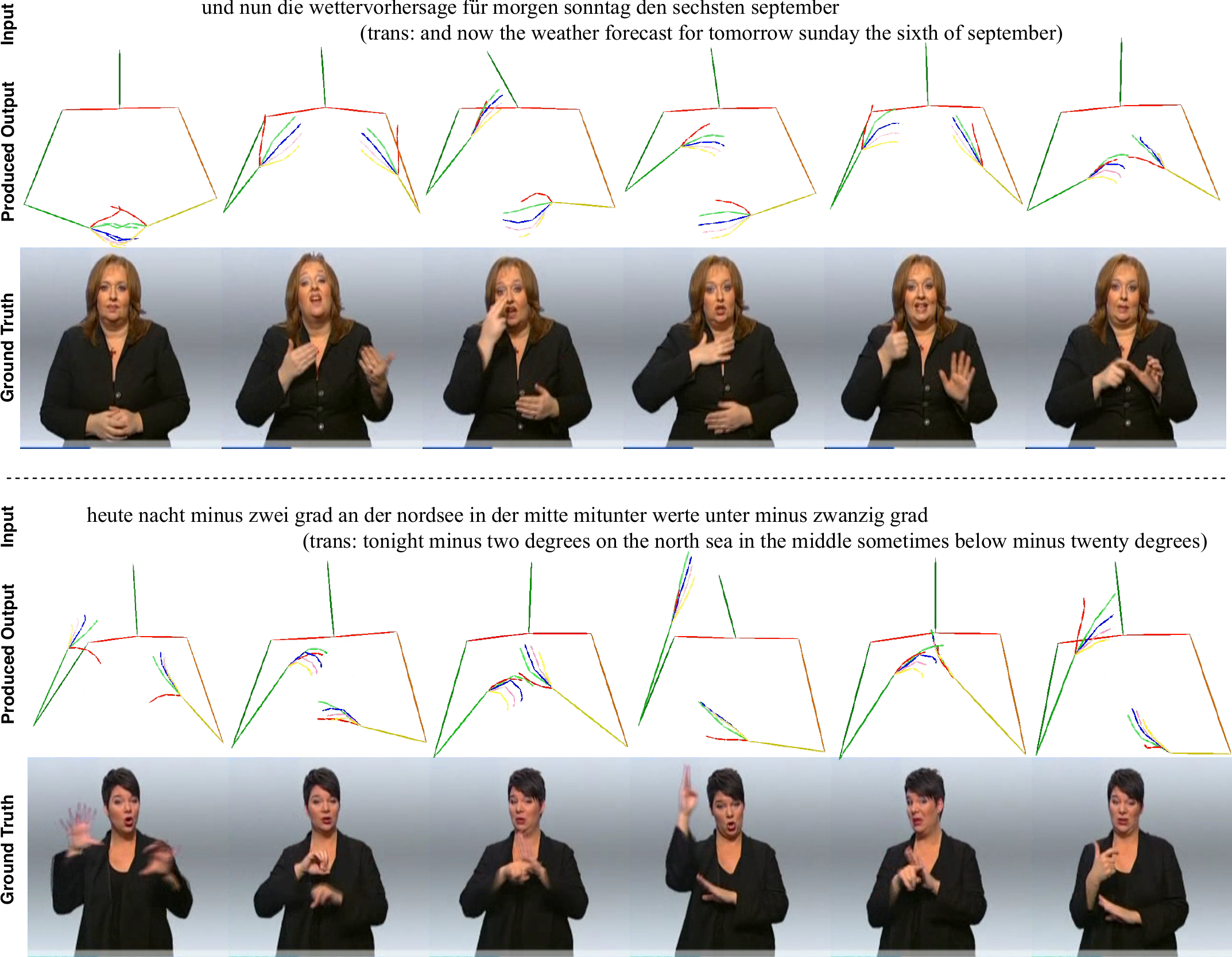}
    \caption{Examples of produced sign pose sequences. The top row shows the spoken language input from the unseen validation set alongside English translation. The middle row presents our produced sign pose sequence from this text input, with the bottom row displaying the ground truth video for comparison.}
    \label{fig:qualitative_output}
\end{figure}%

\section{Qualitative Experiments} \label{sec:qual_experiments}
In this section we report qualitative results for our progressive transformer model. We share snapshot examples of produced sign pose sequences in Figure \ref{fig:qualitative_output}, with more examples provided in supplementary material. The unseen spoken language sequence is shown as input alongside the sign pose sequence produced by our Progressive Transformer model, with ground truth video for comparison.

As can be seen from the provided examples, our \ac{slp} model produces visually pleasing and realistic looking sign with a close correspondence to the ground truth video. Body motion is smooth and accurate, whilst hand shapes are meaningful if a little under-expressed. We find that the most difficult production occurs with proper nouns and specific entities, due to the lack of grammatical context and examples in the training data.

These examples show that regressing continuous sequences can be successfully achieved using an attention-based mechanism. The predicted joint locations for neighbouring frames are closely positioned, showing that the model has learnt the subtle movement of the signer. Smooth transitions between signs are produced, highlighting a difference from the discrete generation of spoken language.

\section{Conclusion} \label{sec:conc}
\acf{slp} is an important task to improve communication between the Deaf and hearing. Previous work has focused on producing concatenated isolated signs instead of full continuous sign language sequences. In this paper, we proposed Progressive Transformers, a novel transformer architecture that can translate from discrete spoken language to continuous sign pose sequences. We introduced a counter decoding that enables continuous sequence generation without the need for an explicit end of sequence token. Two model configurations were presented, an end-to-end network that produces sign direct from text and a stacked network that utilises a gloss intermediary.

We evaluated our approach on the challenging \ac{ph14t} dataset, setting baselines for future research with a back translation evaluation mechanism. Our experiments showed the importance of several data augmentation techniques to reduce model drift and improve \ac{slp} performance. Furthermore, we have shown that a direct text to pose translation configuration can outperform a gloss intermediary model, meaning \ac{slp} models are not limited to only training on data where expensive gloss annotation is available.

As future work, we would like to expand our network to multi-channel sign production, focusing on non-manual aspects of sign language such as body pose, facial expressions and mouthings. It would be interesting to condition a \ac{gan} to produce sign videos, learning a prior for each sign represented in the data.

\section{Acknowledgements}
This work received funding from the SNSF Sinergia project 'SMILE' (CRSII2 160811), the European Union's Horizon2020 research and innovation programme under grant agreement no. 762021 'Content4All' and the EPSRC project 'ExTOL' (EP/R03298X/1). This work reflects only the authors view and the Commission is not responsible for any use that may be made of the information it contains. We would also like to thank NVIDIA Corporation for their GPU grant.

\bibliographystyle{splncs04}
\bibliography{bibliography}

\newpage

\begin{figure}[]
    \includegraphics[width=0.89\linewidth]{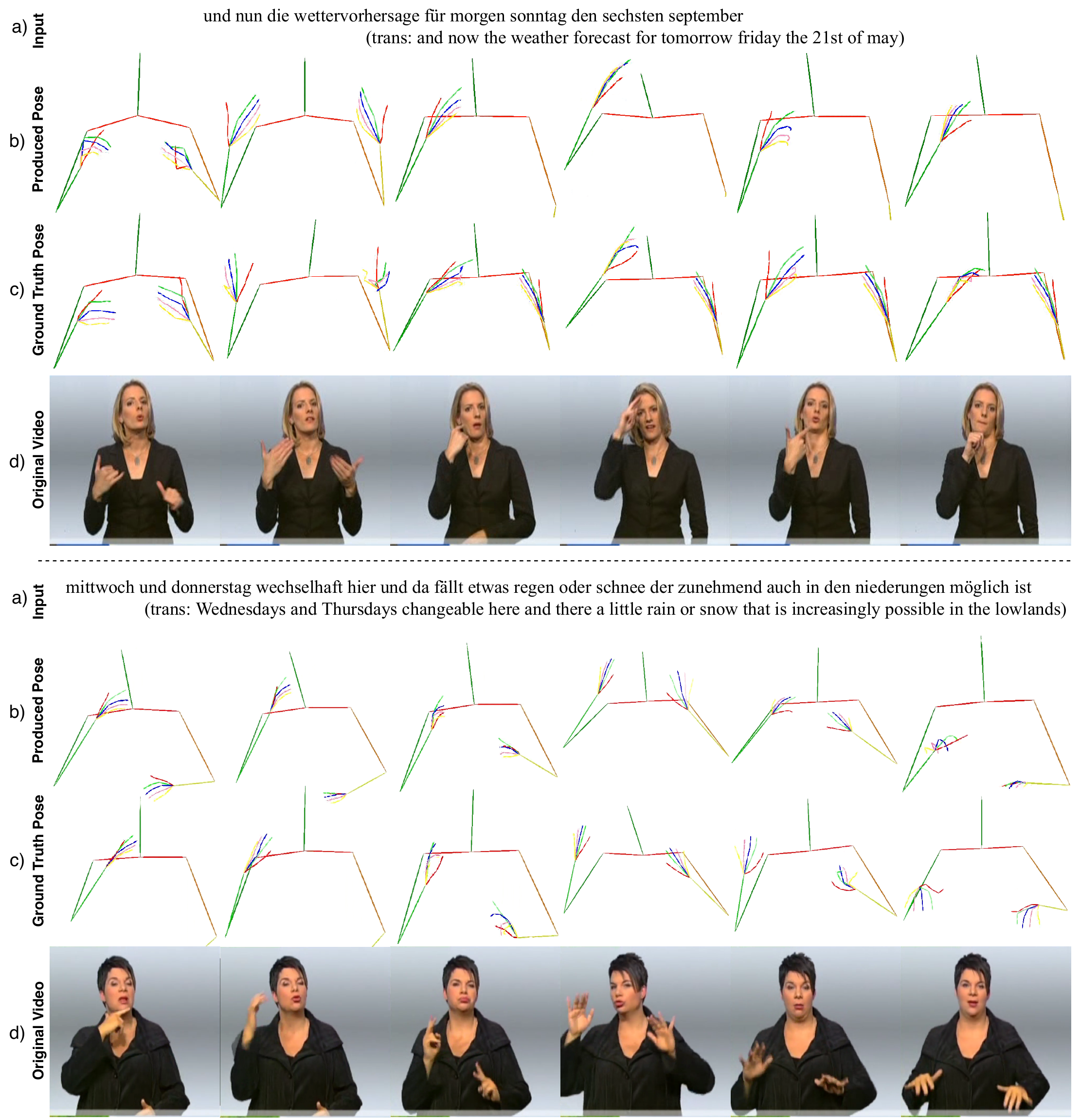}
    \centering
    \caption{Sign language production examples showing (a) spoken language input, (b) produced sign pose, (c) ground truth pose and (d) original video.}
    \label{fig:qualitative_output_1}
\end{figure}

\begin{figure}[]
    \includegraphics[width=0.95\linewidth]{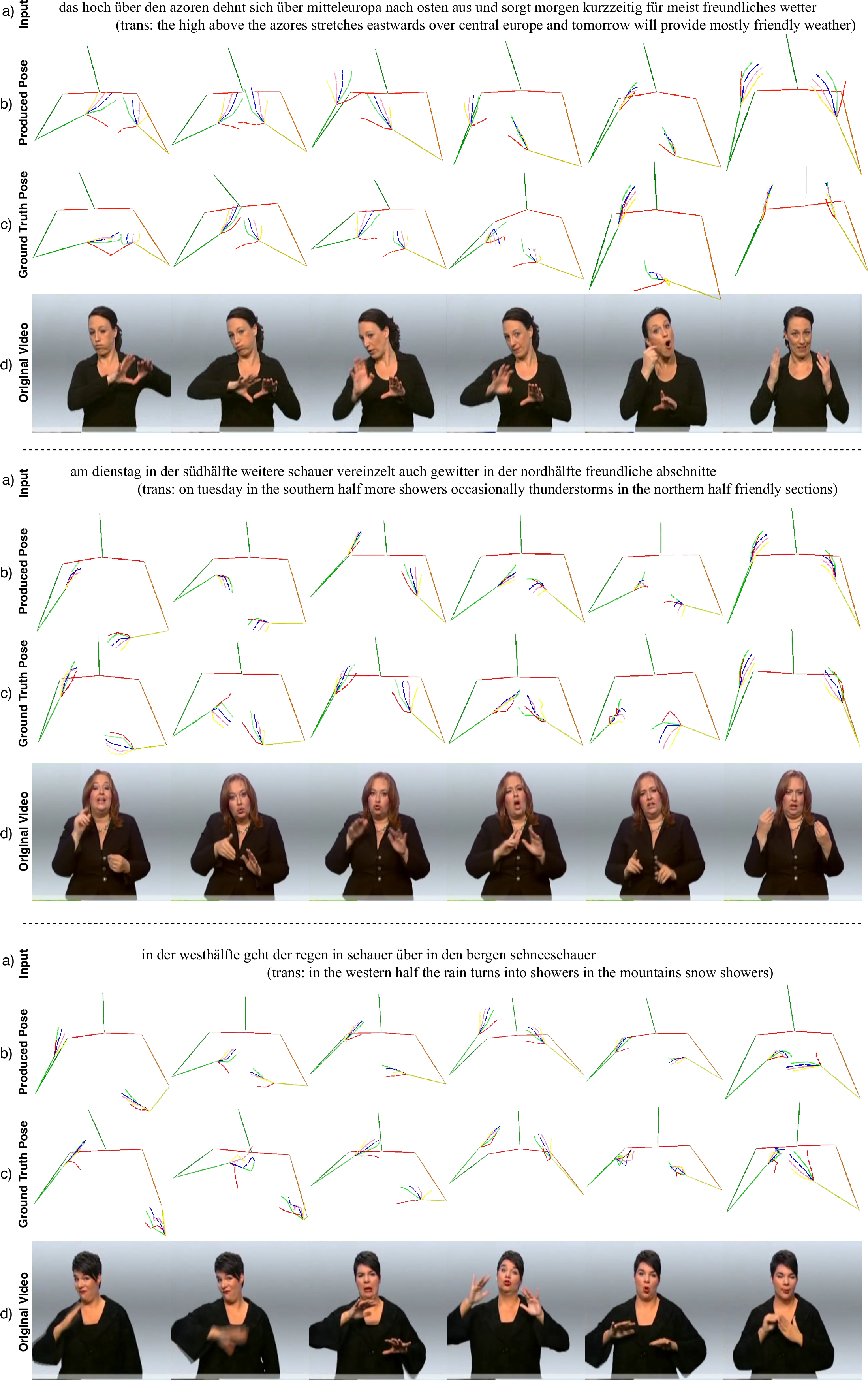}
    \centering
    \caption{Sign language production examples showing (a) spoken language input, (b) produced sign pose, (c) ground truth pose and (d) original video.}
    \label{fig:qualitative_output_2}
\end{figure}

\title{Progressive Transformers for End-to-End Sign Language Production: Supplementary Material}
\titlerunning{Progressive Transformers for SLP}
\author{Ben Saunders\and
Necati Cihan Camgoz \and
Richard Bowden}
\authorrunning{B. Saunders et al.}
\institute{University of Surrey\\
\email{\{b.saunders,n.camgoz,r.bowden\}@surrey.ac.uk}}
\maketitle
In this supplementary material, we give further qualitative results for our Progressive Transformer \acs{slp} model. We share snapshot examples of produced sign pose sequences in Figures \ref{fig:qualitative_output_1} and \ref{fig:qualitative_output_2}, showing (a) input spoken language alongside (b) produced sign pose sequence, with (c) ground truth pose and (d) original video provided for comparison.

\end{document}

%% file: acronyms.tex
\begin{acronym}[PHOENIX14\textbf{T} ]

\acrodefplural{rnn}[RNNs]{Recurrent Neural Networks}
\acrodefplural{cnn}[CNNs]{Convolutional Neural Networks}
\acrodefplural{hmm}[HMMs]{Hidden Markov Models}
\acrodefplural{gru}[GRUs]{Gated Recurrent Units}
\acrodefplural{crf}[CRFs]{Conditional Random Fields}
\acrodefplural{gan}[GANs]{Generative Adversarial Networks}
\acrodefplural{gpu}[GPUs]{Graphic Processing Units}
\acro{bsl}[BSL]{British Sign Language}
\acro{bleu}[BLEU]{Bilingual Evaluation Understudy}
\acro{blstm}[BLSTM]{Bidirectional Long Short-Term Memory}
\acro{cnn}[CNN]{Convolutional Neural Network}
\acro{crf}[CRF]{Conditional Random Field}
\acro{cslr}[CSLR]{Continuous Sign Language Recognition}
\acro{ctc}[CTC]{Connectionist Temporal Classification}
\acro{dl}[DL]{Deep Learning}
\acro{dgs}[DGS]{German Sign Language - Deutsche Gebärdensprache}
\acro{dsgs}[DSGS]{Swiss German Sign Language - Deutschschweizer Geb\"ardensprache}
\acro{dtw}[DTW]{Dynamic Time Warping}
\acro{fc}[FC]{Fully Connected}
\acro{ff}[FF]{Feed Forward}
\acro{gan}[GAN]{Generative Adversarial Network}
\acro{gpu}[GPU]{Graphics Processing Unit}
\acro{gru}[GRU]{Gated Recurrent Unit}
\acro{gtpt}[G2PT]{Gloss to Pose Transformer}
\acro{hmm}[HMM]{Hidden Markov Model}
\acro{isl}[ISL]{Irish Sign Language}
\acro{lstm}[LSTM]{Long Short-Term Memory}
\acro{mha}[MHA]{Multi-Headed Attention}
\acro{mtc}[MTC]{Monocular Total Capture}
\acro{mse}[MSE]{Mean Squared Error}
\acro{nmt}[NMT]{Neural Machine Translation}
\acro{nlp}[NLP]{Natural Language Processing}
\acro{ph12}[PHOENIX12]{RWTH-PHOENIX-Weather-2012}
\acro{ph14}[PHOENIX14]{RWTH-PHOENIX-Weather-2014}
\acro{ph14t}[PHOENIX14\textbf{T}]{RWTH-PHOENIX-Weather-2014\textbf{T}}
\acro{pof}[POF]{Part Orientation Field}
\acro{paf}[PAF]{Part Affinity Field}
\acro{pttt}[P2TT]{Pose to Text Transformer}
\acro{relu}[RELU]{Rectified Linear Units}
\acro{rnn}[RNN]{Recurrent Neural Network}
\acro{rouge}[ROUGE]{Recall-Oriented Understudy for Gisting Evaluation}
\acro{sgd}[SGD]{Stochastic Gradient Descent}
\acro{sla}[SLA]{Sign Language Assessment}
\acro{slr}[SLR]{Sign Language Recognition}
\acro{slt}[SLT]{Sign Language Translation}
\acro{slp}[SLP]{Sign Language Production}
\acro{smt}[SMT]{Statistical Machine Translation}

\acro{ttgt}[T2GT]{Text to Gloss Transformer}
\acro{ttpt}[T2PT]{Text to Pose Transformer}
\acro{ttp}[T2P]{Text to Pose}
\acro{ttgtp}[T2G2P]{Text to Gloss to Pose}
\acro{wer}[WER]{Word Error Rate}

\end{acronym}

%% file: Progressive_Transformers_for_E2E_Sign_Language_Production.bbl
\begin{thebibliography}{10}
\providecommand{\url}[1]{\texttt{#1}}
\providecommand{\urlprefix}{URL }
\providecommand{\doi}[1]{https://doi.org/#1}

\bibitem{ahn2018text2action}
Ahn, H., Ha, T., Choi, Y., Yoo, H., Oh, S.: {Text2Action: Generative
  Adversarial Synthesis from Language to Action}. In: International Conference
  on Robotics and Automation (ICRA) (2018)

\bibitem{ba2016layer}
Ba, J.L., Kiros, J.R., Hinton, G.E.: {Layer Normalization}. arXiv preprint
  arXiv:1607.06450  (2016)

\bibitem{bahdanau2014neural}
Bahdanau, D., Cho, K., Bengio, Y.: {Neural Machine Translation by Jointly
  Learning to Align and Translate}. arXiv:1409.0473  (2014)

\bibitem{bauer2000video}
Bauer, B., Hienz, H., Kraiss, K.F.: {Video-Based Continuous Sign Language
  Recognition using Statistical Methods}. In: Proceedings of 15th International
  Conference on Pattern Recognition (ICPR) (2000)

\bibitem{berndt1994dtw}
Berndt, D.J., Clifford, J.: {Using Dynamic Time Warping to Find Patterns in
  Time Series}. In: AAA1 Workshop on Knowledge Discovery in Databases (KDD)
  (1994)

\bibitem{camgoz2017subunets}
Camgoz, N.C., Hadfield, S., Koller, O., Bowden, R.: {SubUNets: End-to-end Hand
  Shape and Continuous Sign Language Recognition}. In: Proceedings of the IEEE
  International Conference on Computer Vision (ICCV) (2017)

\bibitem{camgoz2018neural}
Camgoz, N.C., Hadfield, S., Koller, O., Ney, H., Bowden, R.: {Neural Sign
  Language Translation}. In: Proceedings of the IEEE Conference on Computer
  Vision and Pattern Recognition (CVPR) (2018)

\bibitem{camgoz2020sign}
Camgoz, N.C., Koller, O., Hadfield, S., Bowden, R.: {Sign Language
  Transformers: Joint End-to-end Sign Language Recognition and Translation}.
  In: Proceedings of the IEEE Conference on Computer Vision and Pattern
  Recognition (CVPR) (2020)

\bibitem{cao2018openpose}
Cao, Z., Hidalgo, G., Simon, T., Wei, S.E., Sheikh, Y.: {OpenPose: Realtime
  Multi-Person 2D Pose Estimation using Part Affinity Fields}. In: Proceedings
  of the IEEE Conference on Computer Vision and Pattern Recognition (CVPR)
  (2017)

\bibitem{cho2014properties}
Cho, K., van Merri{\"e}nboer, B., Bahdanau, D., Bengio, Y.: {On the Properties
  of Neural Machine Translation: Encoder--Decoder Approaches}. In: Proceedings
  of the Syntax, Semantics and Structure in Statistical Translation (SSST)
  (2014)

\bibitem{cooper2012sign}
Cooper, H., Ong, E.J., Pugeault, N., Bowden, R.: {Sign Language Recognition
  using Sub-units}. Journal of Machine Learning Research (JMLR)  \textbf{13}
  (2012)

\bibitem{cui2017recurrent}
Cui, R., Liu, H., Zhang, C.: {Recurrent Convolutional Neural Networks for
  Continuous Sign Language Recognition by Staged Optimization}. In: Proceedings
  of the IEEE Conference on Computer Vision and Pattern Recognition (CVPR)
  (2017)

\bibitem{dai2019transformer}
Dai, Z., Yang, Z., Yang, Y., Carbonell, J., Le, Q.V., Salakhutdinov, R.:
  {Transformer-XL: Attentive Language Models Beyond a Fixed-Length Context}.
  In: International Conference on Learning Representations (ICLR) (2019)

\bibitem{devlin2018bert}
Devlin, J., Chang, M.W., Lee, K., Toutanova, K.: {BERT: Pre-training of Deep
  Bidirectional Transformers for Language Understanding}. In: Proceedings of
  the Conference of the North American Chapter of the Association for
  Computational Linguistics (ACL) (2018)

\bibitem{duarte2019cross}
Duarte, A.C.: {Cross-modal Neural Sign Language Translation}. In: Proceedings
  of the ACM International Conference on Multimedia (ICME) (2019)

\bibitem{ebling2018smile}
Ebling, S., Camg{\"o}z, N.C., Braem, P.B., Tissi, K., Sidler-Miserez, S.,
  Stoll, S., Hadfield, S., Haug, T., Bowden, R., Tornay, S., et~al.: {SMILE:
  Swiss German Sign Language Dataset}. In: Proceedings of the International
  Conference on Language Resources and Evaluation (LREC) (2018)

\bibitem{forster2014extensions}
Forster, J., Schmidt, C., Koller, O., Bellgardt, M., Ney, H.: {Extensions of
  the Sign Language Recognition and Translation Corpus RWTH-PHOENIX-Weather}.
  In: Proceedings of the International Conference on Language Resources and
  Evaluation (LREC) (2014)

\bibitem{ginosar2019learning}
Ginosar, S., Bar, A., Kohavi, G., Chan, C., Owens, A., Malik, J.: {Learning
  Individual Styles of Conversational Gesture}. In: Proceedings of the IEEE
  Conference on Computer Vision and Pattern Recognition (CVPR) (2019)

\bibitem{girdhar2019video}
Girdhar, R., Carreira, J., Doersch, C., Zisserman, A.: {Video Action
  Transformer Network}. In: Proceedings of the IEEE Conference on Computer
  Vision and Pattern Recognition (CVPR) (2019)

\bibitem{glauert2006vanessa}
Glauert, J., Elliott, R., Cox, S., Tryggvason, J., Sheard, M.: {VANESSA: A
  System for Communication between Deaf and Hearing People}. Technology and
  Disability  (2006)

\bibitem{glorot2010understanding}
Glorot, X., Bengio, Y.: {Understanding the Difficulty of Training Deep
  Feedforward Neural Networks}. In: Proceedings of the International Conference
  on Artificial Intelligence and Statistics (AISTATS) (2010)

\bibitem{graves2013generating}
Graves, A.: {Generating Sequences With Recurrent Neural Networks}. arXiv
  preprint arXiv:1308.0850  (2013)

\bibitem{he2016deep}
He, K., Zhang, X., Ren, S., Sun, J.: {Deep Residual Learning for Image
  Recognition}. In: Proceedings of the IEEE Conference on Computer Vision and
  Pattern Recognition (CVPR) (2016)

\bibitem{huang2018music}
Huang, C.Z.A., Vaswani, A., Uszkoreit, J., Shazeer, N., Simon, I., Hawthorne,
  C., Dai, A.M., Hoffman, M.D., Dinculescu, M., Eck, D.: {Music Transformer}.
  In: International Conference on Learning Representations (ICLR) (2018)

\bibitem{isola2017image}
Isola, P., Zhu, J.Y., Zhou, T., Efros, A.A.: {Image-to-Image Translation with
  Conditional Adversarial Networks}. In: Proceedings of the IEEE Conference on
  Computer Vision and Pattern Recognition (CVPR) (2017)

\bibitem{kalchbrenner2013recurrent}
Kalchbrenner, N., Blunsom, P.: {Recurrent Continuous Translation Models}. In:
  Proceedings of the Conference on Empirical Methods in Natural Language
  Processing (EMNLP) (2013)

\bibitem{karpouzis2007educational}
Karpouzis, K., Caridakis, G., Fotinea, S.E., Efthimiou, E.: {Educational
  Resources and Implementation of a Greek Sign Language Synthesis
  Architecture}. Computers \& Education (CAEO)  (2007)

\bibitem{kayahan2019hybrid}
Kayahan, D., G{\"u}ng{\"o}r, T.: {A Hybrid Translation System from Turkish
  Spoken Language to Turkish Sign Language}. In: IEEE International Symposium
  on INnovations in Intelligent SysTems and Applications (INISTA) (2019)

\bibitem{kingma2014adam}
Kingma, D.P., Ba, J.: {Adam: A Method for Stochastic Optimization}. In:
  Proceedings of the International Conference on Learning Representations
  (ICLR) (2014)

\bibitem{kipp2011sign}
Kipp, M., Heloir, A., Nguyen, Q.: {Sign Language Avatars: Animation and
  comprehensibility}. In: International Workshop on Intelligent Virtual Agents
  (IVA) (2011)

\bibitem{ko2019neural}
Ko, S.K., Kim, C.J., Jung, H., Cho, C.: {Neural Sign Language Translation based
  on Human Keypoint Estimation}. Applied Sciences  (2019)

\bibitem{koller2019weakly}
Koller, O., Camgoz, N.C., Bowden, R., Ney, H.: {Weakly Supervised Learning with
  Multi-Stream CNN-LSTM-HMMs to Discover Sequential Parallelism in Sign
  Language Videos}. {IEEE Transactions on Pattern Analysis and Machine
  Intelligence (TPAMI)}  (2019)

\bibitem{koller2015continuous}
Koller, O., Forster, J., Ney, H.: {Continuous Sign Language Recognition:
  Towards Large Vocabulary Statistical Recognition Systems Handling Multiple
  Signers}. Computer Vision and Image Understanding (CVIU)  (2015)

\bibitem{koller2016deephand}
Koller, O., Ney, H., Bowden, R.: {Deep Hand: How to Train a CNN on 1 Million
  Hand Images When Your Data Is Continuous and Weakly Labelled}. In:
  Proceedings of the IEEE Conference on Computer Vision and Pattern Recognition
  (CVPR) (2016)

\bibitem{koller2017resign}
Koller, O., Zargaran, S., Ney, H.: {Re-Sign: Re-Aligned End-to-End Sequence
  Modelling with Deep Recurrent CNN-HMMs}. In: Proceedings of the IEEE
  Conference on Computer Vision and Pattern Recognition (CVPR) (2017)

\bibitem{koller2016deepsign}
Koller, O., Zargaran, S., Ney, H., Bowden, R.: {Deep Sign: Hybrid CNN-HMM for
  Continuous Sign Language Recognition}. In: Proceedings of the British Machine
  Vision Conference (BMVC) (2016)

\bibitem{kouremenos2018statistical}
Kouremenos, D., Ntalianis, K.S., Siolas, G., Stafylopatis, A.: {Statistical
  Machine Translation for Greek to Greek Sign Language Using Parallel Corpora
  Produced via Rule-Based Machine Translation}. In: IEEE 31st International
  Conference on Tools with Artificial Intelligence (ICTAI) (2018)

\bibitem{JoeyNMT}
Kreutzer, J., Bastings, J., Riezler, S.: {Joey {NMT}: A Minimalist {NMT}
  Toolkit for Novices}. In: Proceedings of the Conference on Empirical Methods
  in Natural Language Processing (EMNLP) (2019)

\bibitem{lee2019dancing}
Lee, H.Y., Yang, X., Liu, M.Y., Wang, T.C., Lu, Y.D., Yang, M.H., Kautz, J.:
  {Dancing to Music}. In: Advances in Neural Information Processing Systems
  (NIPS) (2019)

\bibitem{li2019entangled}
Li, G., Zhu, L., Liu, P., Yang, Y.: {Entangled Transformer for Image
  Captioning}. In: Proceedings of the IEEE International Conference on Computer
  Vision (CVPR) (2019)

\bibitem{li2019neural}
Li, N., Liu, S., Liu, Y., Zhao, S., Liu, M.: {Neural Speech Synthesis with
  Transformer Network}. In: Proceedings of the AAAI Conference on Artificial
  Intelligence (2019)

\bibitem{mcdonald2016automated}
McDonald, J., Wolfe, R., Schnepp, J., Hochgesang, J., Jamrozik, D.G., Stumbo,
  M., Berke, L., Bialek, M., Thomas, F.: {Automated Technique for Real-Time
  Production of Lifelike Animations of American Sign Language}. Universal
  Access in the Information Society (UAIS)  (2016)

\bibitem{mikolov2013distributed}
Mikolov, T., Sutskever, I., Chen, K., Corrado, G.S., Dean, J.: {Distributed
  Representations of Words and Phrases and their Compositionality}. In:
  Advances in Neural Information Processing Systems (NIPS) (2013)

\bibitem{mukherjee2019predicting}
Mukherjee, S., Ghosh, S., Ghosh, S., Kumar, P., Roy, P.P.: {Predicting
  Video-frames Using Encoder-convlstm Combination}. In: IEEE International
  Conference on Acoustics, Speech and Signal Processing (ICASSP) (2019)

\bibitem{orbay2020neural}
Orbay, A., Akarun, L.: {Neural Sign Language Translation by Learning
  Tokenization}. arXiv preprint arXiv:2002.00479  (2020)

\bibitem{ozdemir2016isolated}
{\"O}zdemir, O., Camg{\"o}z, N.C., Akarun, L.: {Isolated Sign Language
  Recognition using Improved Dense Trajectories}. In: Proceedings of the Signal
  Processing and Communication Application Conference (SIU) (2016)

\bibitem{parmar2018image}
Parmar, N., Vaswani, A., Uszkoreit, J., Kaiser, {\L}., Shazeer, N., Ku, A.,
  Tran, D.: {Image Transformer}. In: International Conference on Machine
  Learning (ICML) (2018)

\bibitem{paszke2017automatic}
Paszke, A., Gross, S., Chintala, S., Chanan, G., Yang, E., DeVito, Z., Lin, Z.,
  Desmaison, A., Antiga, L., Lerer, A.: {Automatic Differentiation in PyTorch}.
  In: NIPS Autodiff Workshop (2017)

\bibitem{plappert2018learning}
Plappert, M., Mandery, C., Asfour, T.: {Learning a Bidirectional Mapping
  between Human Whole-Body Motion and Natural Language using Deep Recurrent
  Neural Networks}. Robotics and Autonomous Systems  (2018)

\bibitem{ren2019fastspeech}
Ren, Y., Ruan, Y., Tan, X., Qin, T., Zhao, S., Zhao, Z., Liu, T.Y.:
  {FastSpeech: Fast, Robust and Controllable Text to Speech}. In: Advances in
  Neural Information Processing Systems (NIPS) (2019)

\bibitem{salimans2016improved}
Salimans, T., Goodfellow, I., Zaremba, W., Cheung, V., Radford, A., Chen, X.:
  {Improved Techniques for Training GANs}. In: Advances in Neural Information
  Processing Systems (NIPS) (2016)

\bibitem{starner1997real}
Starner, T., Pentland, A.: {Real-time American Sign Language Recognition from
  Video using Hidden Markov Models}. Motion-Based Recognition  (1997)

\bibitem{stoll2018sign}
Stoll, S., Camgoz, N.C., Hadfield, S., Bowden, R.: {Sign Language Production
  using Neural Machine Translation and Generative Adversarial Networks}. In:
  Proceedings of the British Machine Vision Conference (BMVC) (2018)

\bibitem{stoll2020text2sign}
Stoll, S., Camgoz, N.C., Hadfield, S., Bowden, R.: {Text2Sign: Towards Sign
  Language Production using Neural Machine Translation and Generative
  Adversarial Networks}. International Journal of Computer Vision (IJCV)
  (2020)

\bibitem{sutskever2014sequence}
Sutskever, I., Vinyals, O., Le, Q.V.: {Sequence to Sequence Learning with
  Neural Networks}. In: Proceedings of the Advances in Neural Information
  Processing Systems (NIPS) (2014)

\bibitem{suzgun2015hospisign}
S{\"u}zg{\"u}n, M., {\"O}zdemir, H., Camg{\"o}z, N., K{\i}nd{\i}ro{\u{g}}lu,
  A., Ba{\c{s}}aran, D., Togay, C., Akarun, L.: {Hospisign: An Interactive Sign
  Language Platform for Hearing Impaired}. Journal of Naval Sciences and
  Engineering (JNSE)  (2015)

\bibitem{tamura1988recognition}
Tamura, S., Kawasaki, S.: {Recognition of Sign Language Motion Images}. Pattern
  Recognition  (1988)

\bibitem{vaswani2017attention}
Vaswani, A., Shazeer, N., Parmar, N., Uszkoreit, J., Jones, L., Gomez, A.N.,
  Kaiser, {\L}., Polosukhin, I.: {Attention Is All You Need}. In: Advances in
  Neural Information Processing Systems (NIPS) (2017)

\bibitem{vila2018end}
Vila, L.C., Escolano, C., Fonollosa, J.A., Costa-juss{\`a}, M.R.: {End-to-End
  Speech Translation with the Transformer}. In: Advances in Speech and Language
  Technologies for Iberian Languages (IberSPEECH) (2018)

\bibitem{vogler1999parallel}
Vogler, C., Metaxas, D.: {Parallel Midden Markov Models for American Sign
  Language Recognition}. In: Proceedings of the IEEE International Conference
  on Computer Vision (ICCV) (1999)

\bibitem{xiao2020skeleton}
Xiao, Q., Qin, M., Yin, Y.: {Skeleton-based Chinese Sign Language Recognition
  and Generation for Bidirectional Communication between Deaf and Hearing
  People}. In: Neural Networks (2020)

\bibitem{yin2020sign}
Yin, K.: {Sign Language Translation with Transformers}. arXiv preprint
  arXiv:2004.00588  (2020)

\bibitem{zelinka2020neural}
Zelinka, J., Kanis, J.: {Neural Sign Language Synthesis: Words Are Our
  Glosses}. In: The IEEE Winter Conference on Applications of Computer Vision
  (WACV) (2020)

\bibitem{zelinka2019nn}
Zelinka, J., Kanis, J., Salajka, P.: {NN-Based Czech Sign Language Synthesis}.
  In: International Conference on Speech and Computer (SPECOM) (2019)

\bibitem{zhang2019ernie}
Zhang, Z., Han, X., Liu, Z., Jiang, X., Sun, M., Liu, Q.: {ERNIE: Enhanced
  Language Representation with Informative Entities}. In: 57th Annual Meeting
  of the Association for Computational Linguistics (ACL) (2019)

\bibitem{zhou2018end}
Zhou, L., Zhou, Y., Corso, J.J., Socher, R., Xiong, C.: {End-to-End Dense Video
  Captioning with Masked Transformer}. In: Proceedings of the IEEE Conference
  on Computer Vision and Pattern Recognition (CVPR) (2018)

\bibitem{zhu2017unpaired}
Zhu, J.Y., Park, T., Isola, P., Efros, A.A.: {Unpaired Image-to-Image
  Translation using Cycle-Consistent Adversarial Networks}. In: Proceedings of
  the IEEE Conference on Computer Vision and Pattern Recognition (CVPR) (2017)

\end{thebibliography}
